# Face Recognition using Hough Peaks extracted from the significant blocks of the Gradient Image

Arindam Kar[1], Debotosh Bhattacharjee[2], Dipak Kumar Basu[2], Mita Nasipuri[2], Mahantapas Kundu[2]
[1] Indian Statistical Institute, Kolkata-700108, India
[2] Department of Computer Science and Engineering, Jadavpur University, Kolkata- 700032, India
Email: {kgparindamkar@gmail.com, debotosh@indiatimes.com, dipakkbasu@gmail.com, m.nasipuri@cse.jdvu.ac.in, mkundu@cse.jdvu.ac.in}

*Abstract*— **This paper proposes a new technique for automatic face recognition using integrated peaks of the Hough transformed significant blocks of the binary gradient image. In this approach firstly the gradient of an image is calculated and a threshold is set to obtain a binary gradient image, which is less sensitive to noise and illumination changes. Secondly, significant blocks are extracted from the absolute gradient image, to extract pertinent information with the idea of dimension reduction. Finally the best fitted Hough peaks are extracted from the Hough transformed significant blocks for efficient face recognition. Then these Hough peaks are concatenated together, which are used as feature in classification process. The efficiency of the proposed method is demonstrated by the experiment on 1100 images from the FRAV2D face database, 2200 images from the FERET database, where the images vary in pose, expression, illumination and scale and 400 images from the ORL face database, where the images slightly vary in pose. Our method has shown 93.3%, 88.5% and 99% recognition accuracy for the FRAV2D, FERET and the ORL database respectively.**

## I. INTRODUCTION

Automated processing of face images has been receiving increasing attention during the last few decades. Image matching is a fundamental aspect of many problems in computer vision, including object or scene recognition, solving for 3D structure from multiple images, stereo correspondence, and motion tracking. Feature selection is one of the most important steps for detection and classification problems. Good features should be discriminative, robust, easy to compute, and efficient. The Feature selection is specifically important for any face recognition scheme. The more significant facial features such as outline of hair and face, position of eyes, nose and mouth can be preserved by very small number coefficients. The raw pixel values of several image statistics such as colour, gradient and filter responses is the simplest choice for image features, has been used for many years in computer vision, e.g., [1-3]. The underlying motivations for our approach originate from the observation that humans achieve at least a basic level of categorization of faces ''at a glance'' even when images of very low resolution are used. Though human beings can detect and identify faces with no effort, building an automated system that accomplishes such objectives is very challenging. The challenges are even more profound when there are large variations due to illumination conditions, change in pose, facial expression. Of all image analysis of human face, feature extraction is of immense importance.

In order to construct the illumination cone, model based methods also require a set of training images for each subject. Furthermore, because of the complexity of light sources and illumination changes, how to compute and obtain physically implemented lower-dimensional subspaces basis images requires deep investigation.

Edge reflects the discontinuity of intensity distribution in an image. It contains contour, structure and shape information of objects in an image. Cognitive psychological studies indicated that human beings recognize line drawings as quickly and almost as accurately as gray-level pictures. Furthermore, edge is insensitive to illumination changes.

In particular, we propose an approach based on face similarity matching measure using Hough peaks [4] as the feature vector, and showed that this approach produces good recognition results even when less than 5% information of the original gray-scale image is retained after selection of the Hough peaks. Alco in this paper a novel methodology applicable to face matching and fast screening of large facial databases is introduced. The proposed shape comparison method operates on edge maps and derives holistic similarity measures, yet, it does not require the solution of the correspondence problem. The use of edge images is important to introduce robustness against changes in illumination.

There are three main contributions in this paper. Firstly, the gradient of several image statistics is computed for an image of interest, such as the image descriptor. Instead of the joint distribution of the image statistics, we use the gradient change as our edge points for Hough transformation, so the dimension becomes much smaller and hence the computational cost is reduces which is independent of the size of the image. Secondly, applying Hough transformation only on the few blocks of the binary gradient image makes the computationally quite fast. As the block based facial features are compared locally, instead of using a general structure, allows to compare faces in terms of mouth, nose and other features in presence of occlusion. Finally we use the concept of clustering to find the centroid of all the Hough peaks obtained for each block and consider only the nearest two peaks instead of all the peaks. This paper describes image features that have many properties that make them suitable for matching different images of an object. The features are invariant to image scaling and rotation, and partially invariant to change in illumination [5].

The remainder of this paper is organized as follows: Section II describes the extraction of most significant blocks from the binary gradient image of the facial image. Section III details with Hough transformation and selection of the Hough peaks from the selected blocks. Section IV deals with the similarity measure and classification rule. In Section V and VI we assesses the performance of the proposed method on the face recognition task by applying it on the FERET [6], FRAV2D [7] and the ORL face [8] databases and finally by comparing with some of the most popular face recognition schemes.

## II. MAP OF AN IMAGE

### A. Binary Gradient Image

In this section the gradient image $I_G$ of an image (I) is calculated by taking summation over the absolute of the change of pixels in all the eight particular directions i.e., (north (N), northeast (NE), east (E), southeast (SE), south (S), southwest (SW), west (W) and northwest (NW)). The absolute binary gradient image $I_{AG}$ is calculated by considering the pixels of $I_G$ those are greater than the average of the mean and median of $I_G$. The absolute gradient image is calculated as:

$$I_G(x,y) = \begin{bmatrix} abs\{I(x-1,y)-I(x,y)\} + abs\{I(x,y-1)-I(x,y)\} + \\ abs\{I(x+1,y)-I(x,y)\} + abs\{I(x,y+1)-I(x,y)\} + \\ abs\{I(x-1,y+1)-I(x,y)\} + abs\{I(x+1,y-1)-I(x,y)\} + \\ abs\{I(x-1,y-1)-I(x,y)\} + abs\{I(x+1,y+1)-I(x,y)\} \end{bmatrix}$$

After calculating the gradient image, edge and Sobel operators [9] are used to calculate the threshold value, which in turn give a binary gradient image. The original image and its corresponding binary gradient image are shown in Fig. 1.(a) and Fig. 1. (b) respectively:

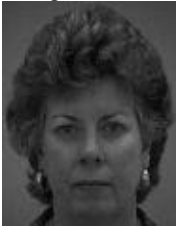 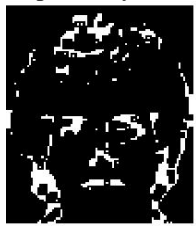

Fig.1. (a)        Fig.1. (b)
Fig 1.(a) Original image, (b) Binary absolute gradient image

The processed binary gradient mask images still shows lines of high contrast in the image. So by using linear structuring elements i.e. by dilation of the binary gradient image, these linear gaps are removed. This is very similar to the way by which human perceive the face of other human being. From figure 1(b) it is clear that the white portions of the binary gradient image, where the change in contrast is very high, and are the region of interest except those boundaries i.e., we are interested to extract those regions that contains the important facial features like nose, mouth, eye etc.

### B. Extraction of Significant Blocks

The step for extracting the significant blocks from the image is given below.
Blocks are generated randomly inside the absolute gradient image $I_G$ and are selected if;
 a) The count of white pixels in the block is greater than M % of the total pixels in the block, where M is the mean of the pixel values of the binary gradient image.
b) If there is any overlap between a selected block and a new block, then that particular block is selected for which the count of white pixels is more.
The extracted significant $k^{th}$ block for the $i^{th}$ training image is defined as $B_{i,j} = \{(x_k, y_k), B_{i,j}(x_k, y_k)\}$. The first two components represent the starting location of the extracted block. These two components of feature are very important during matching (comparison) process. The remaining components are the elements of the block.
d) This process of random block generation and extraction is repeated for X times (say X=500000), and the most significant blocks are selected from the generated blocks in such a way that the blocks captures the important facial features like nose, mouth, eye brows, eyes etc and are in the range of 20-30% of the total blocks in the image. The proposed gradient based method reduces the number of edge pixels by as much as 70-80%.

## III. SIGNIFICANT BLOCK EXTRACTION

### A. Hough tranformation

The Hough transform is widely used for shape analysis in machine vision with its robustness to noise and high efficiency. In this paper, we focus on the standard Hough transform (SHT) [10], which is used to detect straight lines. The SHT uses the parametric representation of a line:

$\rho = x\cos\theta + y\sin\theta$ ,

here ρ is the distance from the origin to the line along a vector perpendicular to the line. θ is the angle of the perpendicular projection from the origin to the line measured in degrees clockwise from the positive x-axis. The range of θ is $-90° \leq \theta \leq 90°$. The angle of the line itself is $\theta + 90°$ measured clockwise with respect to the positive x-axis.

where ρ is the length of the perpendicular of a line passing through (x,y) and subtending an angle θ with the x-axis. Since the point can be defined by a set of intersecting lines that subtend angles between 0 to 90°, a sinusoidal locus is generated using $\rho = x\cos\theta + y\sin\theta$ , for every edge point. A straight line can be characterized by a set of collinear edge points. So each of these edge points have one line in common, which implies that at one unique (ρ,θ) coordinate in Hough space the sinusoids of these points intersect. This is illustrated in Fig. 2.

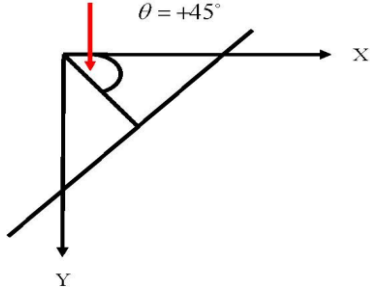

Fig. 2 Representation of Hough transformation

The SHT is a parameter space matrix whose rows and columns correspond to ρ and θ values respectively. The elements in the SHT represent accumulator cells. Peak values in the SHT represent potential lines in the input image. The steps for extracting of Hough peaks from the significant blocks are given below.

a. For each such block $B_i$, i=1, 2,…n Hough transformation is applied.
b. Let the size of each Block be m×m. Firstly consider the top m peaks in decreasing order of their value.
c. Then find the centroid of these m peaks using the K-means cluster algorithm [11], and select only those 2 peaks which are nearest to the centroid as features. Thus we obtain a feature vector of size $2 \times n$, where n is the number of blocks. The proposed selection of peaks has been adjusted experimentally.

Fig. 3(a), (b) and (c) show the selection of the significant block from the binary gradient image, the corresponding block in the original and the selected 2 Hough peaks from the significant block from the FERET dataset, respectively.

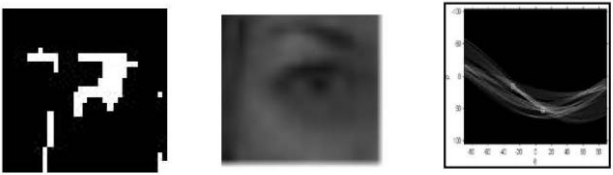

Fig. 3. (a)      Fig. 3. (b)      Fig. 3. (c)

Fig 3. (a) Significant Block extracted from the binary gradient image, (b) Same position block from the original image (c) selected 2 Hough peaks from a significant block

Thus from each block the obtained Hough peaks are concatenated to obtain a feature vector of size 2×n or 2n. This final feature obtained from the Hough peaks are points with high information content of the face image.

## IV. SIMILARITY MEASURE AND CLASSIFICATION

The dissimilarity measure between the $i^{th}$ testing and $j^{th}$ training images is measured as:

$$D(i,j) = \frac{1}{B_i} \sum_{k=1}^{B_i} \left[ \max_{1 \leq l \leq B_j} \left\{ I(k,l). \chi^2(f_{ik}, f_{jl}) \right\} \right],$$

where $B_i$ = number of blocks in the $i^{th}$ testing image, and $B_j$ = number of blocks in the $j^{th}$ training image. Here $I(k,l)$ is defined in such a way that blocks of the training image which are not close enough to a block of the test image in terms of location are discarded, and the extracted features (Hough peaks) of those training blocks which are in the neighborhood of the testing blocks are only compared for finding the similarity measure i.e,

$$I(k,l) = \begin{cases} 1 & if \ \sqrt{(x_k - x_l)^2 + (y_k - y_l)^2} < th1 \\ 0 & otherwise \end{cases},$$

where $th1$ is the threshold. In this experiment the threshold is taken as the block size i.e. the approximate radius of the region that contains the eye. This technique helps to avoid the matching of a feature of a block located around the eye with a point of a training facial image that is located around the mouth. Also $\chi^2(f_1, f_2)$ is the Chi-square [12] dissimilarity measure between vector $f_1$ and $f_2$, and is defined as:

$$\chi^2(f_1, f_2) = \sum_{m=1}^{2} \frac{(f_{1m} - f_{2m})^2}{(f_{1m} + f_{2m})^2},$$ where m is the number

of features (Hough peaks) in the block. Finally, the $i^{th}$ testing image is classified to the $k^{th}$ training image class if $D(i,j) = \min_{j} (D(i,j))$.

## V. EXPERIMENT

The effectiveness of the proposed method has been successfully tested on face recognition using three databases, a) the whole ORL facial database, b) the FRAV2D database containing 1100 frontal face images corresponding to 100 subjects, c) The FERET database, containing 2200 frontal face images corresponding to 200 subjects, which are acquired under variable illumination and facial expression. The effectiveness of the method is shown in terms of both absolute performance indices and comparative performance against some popular face recognition schemes such as the PCA, LDA [13, 14], Gabor wavelets (GW) [15], edge based features, direct Hough transformation, and sub peaks of the Hough transformation [16].

*A. Experiment on the ORL database*

The whole ORL database is considered here. In the experiment each image is scaled to $92 \times 112$ with 256 gray levels. . Fig. 4 shows all samples of one individual. First two images of each individual are considered for training and the remaining images are used as testing samples.

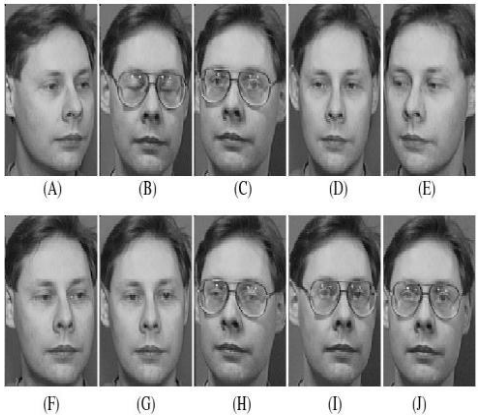

Fig. 4. Demonstration images of one subject from the ORL database

### B. Experiment on the FRAV2D database

The FRAV2D face database, employed in the experiment, consists of 1100 colour face images of 100 individuals, 11 images of each individual are taken, including frontal views of faces with different facial expressions, under different lighting conditions. All colour images are transformed into gray images and scaled to 92×112. Fig. 5 shows all samples of one individual. The details of the images are as follows: (A) regular facial status; (B) and (C) are images with a 15° turn with respect to the camera axis; (D) and (E) are images with a 30° turn with respect to the camera axis; (F) and (G) are images with gestures; (H) and (I) are images with occluded face features; (J) and (K) are images with change of illumination.

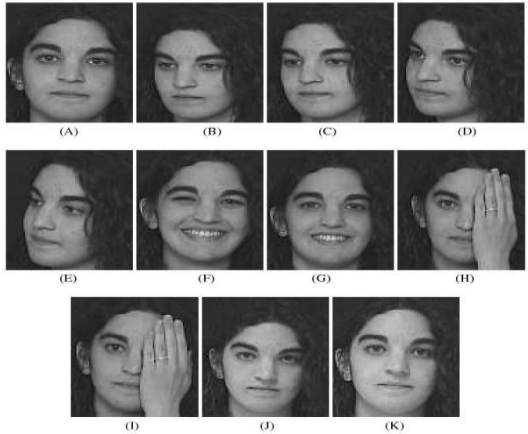

Fig 5. Demonstration images of one subject from the FRAV2D database

### C. Experiment on the FERET database

The FERET database, employed in the experiment here, contains 2,200 facial images corresponding to 200 individuals with each individual contributing 11 images. The images in this database were captured under various illuminations that display, a variety of facial expressions and poses. As the images include the background and the body chest region, so each image is cropped to exclude those, and then scaled to 92 × 112. Fig. 6 shows all samples of one individual. The details of the images are as follows: (A) regular facial status; (B) +15° pose angle; (C) -15° pose angle; (D) +25° pose angle; (E) -25° pose angle; (F) +40° angle; (G) -40° pose angle; (H) +60° pose angle; (I) -60° pose angle; (J) alternative expression; (K) different illumination.

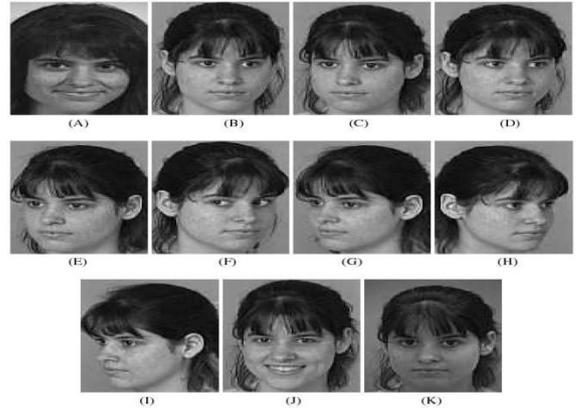

Fig. 6. Demonstration images of an individual from the FERET database

### D. Specificity and Sensitivity measure for the FRAV2D, FERET and the ORL dataset:

To measure the sensitivity and specificity [17] the dataset from the ORL, FRAV2D, & FERET database is prepared in the following manner. In the FRAV2D database for each individual a single class is constituted with 18 images in it. Thus, a total 100 class is obtained from the dataset of 1100 images of 100 individuals. Out of the 18 images in each class, 11 images are of a particular individual, and 7 images are of other individuals taken by permutation. Similarly 200 classes are obtained for the FERET dataset, each class having 18 images in it. Out of the 18 images in each class, 11 images are of a particular individual, and 7 images are of other individuals taken by permutation. For ORL database 40 classes are constructed with 15 images in each class. Out of the 15 images in each class, 10 images are of a particular individual, and 5 images are of other individuals. Using these datasets the true positive ($T_P$); false positive ($F_P$); true negative ($T_N$); false negative ($F_N$); are measured. For all the datasets, the first 2 images (A-B), of a particular individual are selected as training samples and the remaining images of that particular individual are used as positive testing samples. The negative testing is done using the images of the other individuals.

Table I: Specificity and Sensitivity measure of the FERET & FRAV2D dataset:

| Total no. of classes=100, Total no. of images= 1800 ||||
|---|---|---|---|
| FRAV2D | | Individual belonging to a particular class ||
| | | Using first 2 images of an individual as training images ||
| | | Positive | Negative |
| FRAV2D test | Positive | $T_P$ =840 | $F_P$ =7 |
| | Negative | $F_N$ =60 | $T_N$ =693 |
| | | Sensitivity = $T_P$ / ($T_P + F_N$) ≈ 93.33% | Specificity = $T_N$ / ($F_P + T_N$) ≈ 99.3% |

| Total no. of classes=200, Total no. of images= 3600 | | | |
|---|---|---|---|
| FERET | | Using first 2 images of an individual as training images | |
| | | Positive | Negative |
| FERET test | Positive | $T_P$ = 1593 | $F_P$ =14 |
| | Negative | $F_N$ =207 | $T_N$ =1386 |
| | | Sensitivity = $T_P$ / ($T_P + F_N$) ≈ 88.5% | Specificity = $T_N$ / ($F_P + T_N$)=99% |
| Total no. of classes=40, Total no. of images= 600 | | | |
| ORL test | | Using first 2 images of an individual as training images | |
| | Positive | $T_P$ =316 | $F_P$ =0 |
| | Negative | $F_N$ =4 | $T_N$ =200 |
| | | Sensitivity = $T_P$ / ($T_P + F_N$) ≈ 99% | Specificity = $T_N$ / ($F_P + T_N$)=100% |

Thus for **FRAV2D database** considering the first 2 images (A-B) of a particular individual for training the achieved rates are:
False positive rate = FP / (FP + TN) = 100% − Specificity =.7%
False negative rate = FN / (TP + FN) = 100%− Sensitivity=6.67%
**Accuracy = ($T_P$+$T_N$)/($T_P$+$T_N$+$F_P$+$F_N$) ≈ 96.3%.**
For **FERET database** considering the first 2 images (A-B) of a particular individual for training the achieved rates are:
False positive rate = FP / (FP + TN) = 100% − Specificity =1%
False negative rate = FN/(TP+ FN) = 100% − Sensitivity=11.5%
**Accuracy = ($T_P$+$T_N$)/($T_P$+$T_N$+$F_P$+$F_N$) = 93.75%.**
For **ORL database** considering the first 2 images (A-B) of a particular individual for training the achieved rates are:
False positive rate = FP / (FP + TN) = 100% − Specificity =0%
False negative rate = FN/(TP+ FN) = 100% − Sensitivity=**1%**
**Accuracy = ($T_P$+$T_N$)/($T_P$+$T_N$+$F_P$+$F_N$) = 99.5%.**

## VI. RESULTS

Experimental results indicate that a) the extracted Hough peaks as features, from the proposed significant blocks of the absolute gradient image integrated together used for classification using the $\chi^2$ distance similarity measure achieved a verification rate which is as good as any previously used methods like PCA, LDA, Gabor wavelets for face recognition .It is also seen that significant block based binary gradient based feature extraction technique achieves the best accuracy, when peaks are selected from the Hough transformed blocks.; b) During simulations, it is observed that the locations of significant blocks , found from the absolute binary gradient image of the face image, can give small deviations between different conditions (expression, illumination, having glasses or not, rotation, etc.), for the same individual. Therefore, an exact measurement of corresponding distances is not possible unlike the geometrical feature based methods; c) The computational time is significantly reduced by using only few significant blocks of the image, although the accuracy is not compromised; and finally d) selected of only those peaks from the Hough transformed block from the accumulation of peaks which are nearest to the centroid of the available peaks, and hence reduce the size of the feature vector. Extensive experiments indicate that the proposed method has achieved a much better performance than the previous variations of Hough transform. Further it also decreases the effect of occluded features. Table II shows the upper bound performances on the FERET, FRAV2D and ORL facial databases, which reflects that our proposed method achieves higher performance results. Thus the proposed algorithm deals with two of these problems, namely occlusion and illumination changes. Experimental results on all the three databases taking first two images (A-B) as training face and the remaining images as testing images are shown below.

Table II. Performance results of well known face recognition algorithms together with the proposed method on FERET, ORL and FRAV2D respectively with the use of $\chi^2$ similarity measure.

| Databases | Experiment conducted for feature Selection | Similarity Measure | Highest Recognition Accuracy (%) |
|---|---|---|---|
| FERET, FRAV2D, ORL, databases | Significant Blocks of the Binary Gradient Image + Hough Transformation+ Selection of the 2 nearest to the centroid peaks. | $\chi^2$ Distance measure | **88.5, 93.3, 99** |
| FERET, FRAV2D, ORL, databases | Significant Blocks of the Binary Gradient Image + Hough Transformation+ Selection of the all the m peaks, [m= Block size]. | $\chi^2$ Distance measure | **85,88.8, 97** |
| FERET, FRAV2D, ORL, databases | Selecting all the Blocks of the Binary Gradient Image + Hough Transformation+ Selection of the 2 nearest to the centroid peaks. | $\chi^2$ Distance measure | **83.5, 88, 95** |
| FERET, FRAV2D, ORL, databases | Selecting all the Blocks of the Binary Gradient Image + Hough Transformation + Selection of the all the m peaks, [m=Block size]. | $\chi^2$ Distance measure | **84,88,96** |
| FERET, FRAV2D, ORL, databases | Binary Gradient Image (without block breaking)+ Hough Transformation + Selection of all the available peaks. | $\chi^2$ Distance measure | **75.67 ,80.5, 95** |
| FERET, FRAV2D, ORL, databases | Significant Blocks of the Binary Gradient Image | $\chi^2$ Distance measure | **80, 83.8, 97** |
| FERET, FRAV2D, ORL, databases | Sub Peaks of Hough transformation | $\chi^2$ Distance measure | **85., 91,99** |
| FERET, FRAV2D, ORL, databases | All the edge points of the Binary Gradient Image | $\chi^2$ Distance measure | **76.8,86, 97** |
| FERET, FRAV2D, ORL, databases | Gabor Wavelets | $\chi^2$ Distance measure | **79.5,85.4, 96.7** |
| FERET, FRAV2D, ORL, databases | LDA | $\chi^2$ Distance measure | **74.6,86.7,98** |
| FERET, FRAV2D, ORL, databases | PCA | $\chi^2$ Distance measure | **71.5,82.8,97.5** |

## Conclusion

The proposed technique as presented here obtains the absolute gradient of the original image in eight directions. From these gradient images only the informative significant blocks are extracted with our extraction technique, and are used, thus the number of edge points are drastically reduced which alleviate the computational and storage load, for real time applications. As only the significant blocks are used instead of the whole image, so in this approach the facial features are compared locally instead of a general structure, and hence allow us to make decision from the different parts of a face. As such it performs better in presence of occlusions. Again by selection of only those two peaks of the Hough transformed significant blocks that are nearest to the centroid of the available peaks further reduces the dimension of the feature vector and also enhances face recognition in presence of illumination and expression changes as a property of Hough transformation.

## ACKNOWLEDGEMENT

Authors are thankful to a major project entitled "Design and Development of Facial Thermogram Technology for Biometric Security System," funded by University Grants Commission (UGC),India and "DST-PURSE Programme" and CMATER and SRUVM project at Department of Computer Science and Engineering, Jadavpur University, India for providing necessary infrastructure to conduct experiments relating to this work. Dr. D.K. Basu would also like to thank the AICTE, New Delhi, INDIA for providing him an Emeritus Fellowship (F.No.1-51/RID/EF (13)/2007-2008).